\documentclass[10pt,twocolumn,letterpaper]{article}

\PassOptionsToPackage{table,dvipsnames}{xcolor}
\usepackage{cvpr}              
\usepackage{multirow}

\definecolor{cvprblue}{rgb}{0.21,0.49,0.74}
\usepackage[pagebackref,breaklinks,colorlinks,allcolors=cvprblue]{hyperref}
\usepackage{tabularx}
\usepackage{array}
\definecolor{lgray}{RGB}{230,230,230}
\definecolor{lblue}{RGB}{52, 216, 235}
\newcolumntype{L}[1]{>{\raggedright\let\newline\\\arraybackslash\hspace{0pt}}m{#1}}
\newcolumntype{C}[1]{>{\centering\let\newline\\\arraybackslash\hspace{0pt}}m{#1}}
\newcolumntype{R}[1]{>{\raggedleft\let\newline\\\arraybackslash\hspace{0pt}}m{#1}}

\def\paperID{3862} 
\def\confName{CVPR}
\def\confYear{2025}

\title{FSHNet: Fully Sparse Hybrid Network for 3D Object Detection}

\author{
    Shuai~Liu\textsuperscript{1},
    Mingyue~Cui\textsuperscript{1},
    Boyang~Li\textsuperscript{1,}\textsuperscript{2,}\thanks{Corresponding author.}~,
    Quanmin~Liang\textsuperscript{1},
    Tinghe~Hong\textsuperscript{1},
    Kai~Huang\textsuperscript{1}\\
    Yunxiao~Shan\textsuperscript{3},
    Kai~Huang\textsuperscript{1}\\
    \textsuperscript{1}School of Computer Science and Engineering, Sun Yat-sen University \\
    \textsuperscript{2}Key Laboratory of Machine Intelligence and Advanced Computing, Sun Yat-sen University \\
    \textsuperscript{3}School of Artificial Intelligence, Sun Yat-sen University \\
    \texttt{\{liush376@mail2, liby83@mail\}.sysu.edu.cn}
}

\begin{document}
\maketitle

\begin{abstract}
Fully sparse 3D detectors have recently gained significant attention due to their efficiency in long-range detection. However, sparse 3D detectors extract features only from non-empty voxels, which impairs long-range interactions and causes the center feature missing. The former weakens the feature extraction capability, while the latter hinders network optimization. To address these challenges, we introduce the Fully Sparse Hybrid Network (FSHNet). FSHNet incorporates a proposed SlotFormer block to enhance the long-range feature extraction capability of existing sparse encoders. The SlotFormer divides sparse voxels using a slot partition approach, which, compared to traditional window partition, provides a larger receptive field. Additionally, we propose a dynamic sparse label assignment strategy to deeply optimize the network by providing more high-quality positive samples. To further enhance performance, we introduce a sparse upsampling module to refine downsampled voxels, preserving fine-grained details crucial for detecting small objects. Extensive experiments on the Waymo, nuScenes, and Argoverse2 benchmarks demonstrate the effectiveness of FSHNet. The code is available at \url{https://github.com/Say2L/FSHNet}.

\end{abstract}

\section{Introduction}
3D object detection has attracted increasing research interest with its growing range of applications in areas such as autonomous vehicle perception and precision robotic operations. Currently, 3D LiDAR-based object detectors \cite{second, dsvt, dcdet} typically transform sparse features into 2D dense feature maps to leverage established 2D object detection techniques. However, these detectors often waste significant computational and memory resources processing unoccupied voxels. This issue becomes more pronounced when performing long-range detection \cite{fsd}. Therefore, developing fully sparse 3D detectors (abbreviated as sparse detectors) that extract sparse features directly from sparse inputs without dense operations is urgently needed. 

Sparse detectors are still emerging compared to their dense counterparts. Some methods \cite{pointrcnn, 3dssd, iassd} utilize the PointNet series \cite{pointnet, pointnet++} to extract sparse features from raw point clouds. Point R-CNN \cite{pointrcnn} generates proposals for foreground points, 3DSSD \cite{3dssd} improves speed by eliminating the upsampling layer in PointNet, and IA-SSD \cite{iassd} trims background points to further enhance efficiency. Other methods \cite{fsd, voxelnext, safdnet} convert points into voxels for feature extraction and prediction. FSD \cite{fsd} employs instance-level optimization to preserve essential features, VoxelNeXt \cite{voxelnext} predicts on the nearest voxels to object centers, and SAFDNet \cite{safdnet} mitigates the center feature missing through adaptive feature diffusion. Currently, there is no standardized solution to build sparse detectors with robust feature extraction capability and superior optimization.


Constructing a robust and highly optimized sparse detector presents two key challenges. First, establishing long-range interactions is difficult. Sparse detectors \cite{fsd, voxelnext, safdnet} typically operate only on non-empty voxels for efficiency, which often results in the inability to maintain connections between distant elements. As shown in Figure~\ref{fig1}, two voxels become isolated because they lie outside the range of the convolution kernel. This issue does not affect dense detectors, where dense convolutions can establish connections between distant voxels through intermediary ones. Second, network optimization is challenging. Since non-empty voxels generally occupy a small proportion of each object, the object centers are often empty, particularly in the case of large objects. Many popular detectors \cite{centerpoint, afdet, dsvt} have shown that the center feature serves as an effective proxy for an object. As a result, the absence of center features hampers the network optimization \cite{fsd}.

To address the challenges outlined above, we introduce a sparse detector called the Fully Sparse Hybrid Network (FSHNet). Specifically, to tackle the first challenge, FSHNet incorporates a SlotFormer block to enable long-range interactions between sparse voxels. Unlike traditional square windows, SlotFormer divides sparse voxels into several `slots' where each slot represents a window with infinite edge length on one side. The block then uses a linear attention mechanism \cite{performer, efficientvit, han2023flatten, linear_attn} to establish connections between voxels within each slot. In different layers of SlotFormer, we alternate the direction of the slots to facilitate global-range interactions. By combining SlotFormer with an existing sparse convolution encoder \cite{voxelnext, safdnet}, we overcome the limitations of current encoders in handling long-range interactions. To address the second challenge, i.e., network optimization, we propose a dynamic sparse label assignment method that dynamically selects positive samples from several nearest voxels to each object center, providing high-quality positive samples for training. Additionally, we introduce a sparse upsampling module to refine the sparse voxels, thus retaining fine-grained features which is useful for detecting small objects. 

We compare our FSHNet with current state-of-the-art sparse detectors \cite{fsd, voxelnext, safdnet} and advanced dense detectors \cite{dsvt, dcdet, scatterformer}. Experiments are conducted on the Waymo Open \cite{waymo}, nuScenes \cite{nuscenes}, and Argoverse2 \cite{argo2} datasets. The results show that FSHNet significantly outperforms existing methods. The contributions of this work can be summarized as follows:

\begin{itemize}
    \item We propose FSHNet, a 3D sparse detector that combines the efficiency of sparse convolution with the long-range interaction capability of the attention mechanism in a single model.
    \item We introduce the SlotFormer block to establish global-range interactions between sparse voxels.
    \item We design a dynamic sparse label assignment strategy to improve the optimization of the sparse detector.
    \item Extensive experiments on the Waymo Open \cite{waymo}, nuScenes \cite{nuscenes}, and Argoverse2 \cite{argo2} datasets demonstrate the effectiveness and generalization of our method.
\end{itemize}

\section{Related Work}
\subsection{LiDAR-based 3D Object Detection}
LiDAR-based 3D object detectors can be categorized into dense detectors \cite{voxelnet, second, pointpillars, voxel_rcnn, hednet} and sparse detectors \cite{fsd, swformer, rsn, voxelnext, safdnet}. Dense detectors convert sparse voxel feature maps into dense feature maps, using dense convolution layers and dense detection heads to produce predictions similar to those in 2D object detectors \cite{faster_rcnn, mask_rcnn}. In contrast, sparse detectors leverage the inherent sparsity of point clouds by employing fully sparse networks and sparse detection heads for prediction.

\noindent\textbf{Dense Detectors. }
VoxelNet \cite{voxelnet} encodes voxel features using PointNet \cite{pointnet} and then extracts features from 3D feature maps through 3D convolutions. SECOND \cite{second} improves the efficiency of encoding sparse voxel features by introducing 3D sparse convolution. PointPillars \cite{pointpillars} divides the point cloud into pillar voxels, eliminating the need for 3D convolution and enabling high inference speed. Voxel R-CNN \cite{voxel_rcnn} replaces raw point features with 3D voxel features within the 3D backbone for second-stage refinement. HEDNet \cite{hednet} employs encoder-decoder blocks to capture long-range dependencies among voxel features.

\noindent\textbf{Sparse Detectors. }RSN \cite{rsn} performs foreground segmentation on range images, and then generates box predictions from sparse foreground points. FSD \cite{fsd} uses an instance-level feature extraction module to mitigate the issue of center feature missing in sparse detectors. SWFormer \cite{swformer} introduces sparse window transformers, creating a fully transformer-based network for 3D object detection. VoxelNeXt \cite{voxelnext} directly predicts objects from sparse voxel features, eliminating the need for hand-crafted proxies. SAFDNet \cite{safdnet} incorporates an adaptive feature diffusion strategy to address the center feature missing issue.

\begin{figure}[t]
    \centering
    \includegraphics[width=0.8\columnwidth]{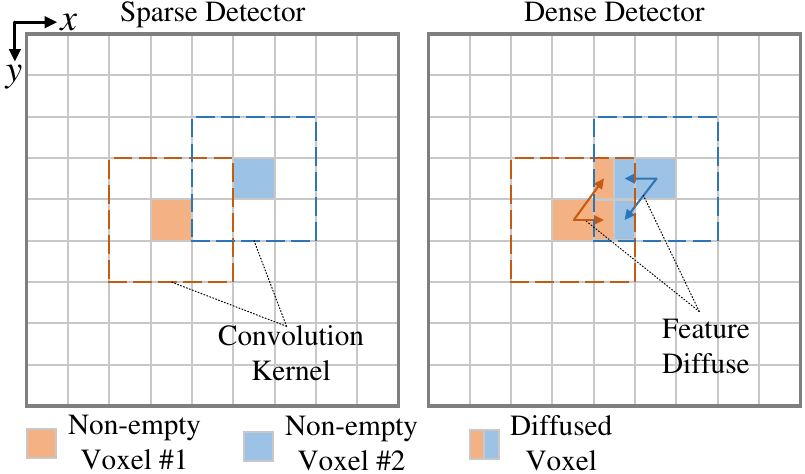}
    \caption{Comparison of voxel interactions between sparse and dense detectors. Left: two distant voxels do not interact because they fall outside each other's convolution kernel range. Right: interaction occurs between distant voxels through a diffused voxel located between them. }
    \label{fig1}
\end{figure}

\begin{figure*}[t]
    \centering
    \includegraphics[width=0.9\textwidth]{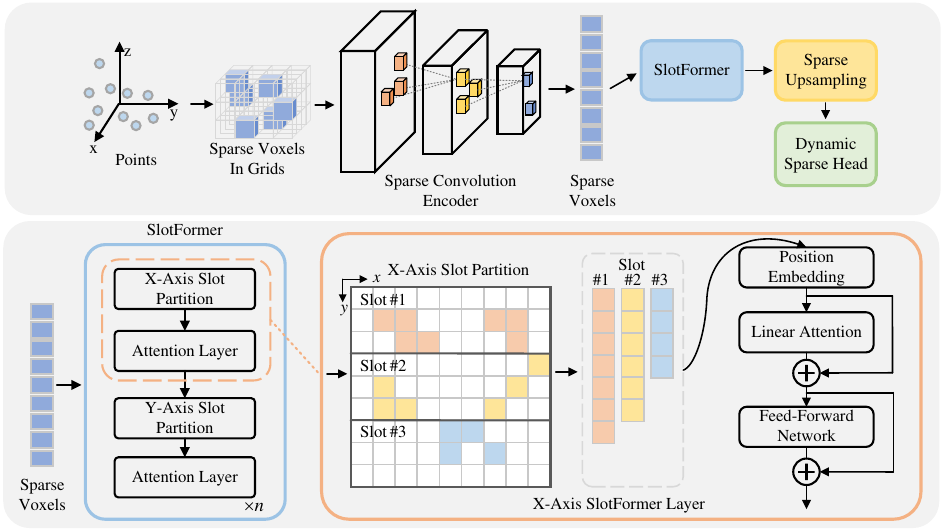}
    \caption{The overall framework of our FSHNet. Taking the raw point clouds as input, FSHNet first transforms points to voxels, and then it employs the sparse convolution encoder, SlotFormer, and sparse upsampling module to extract high-level and fine-grained sparse features. Finally, the dynamic sparse head is used to generate predictions. }
    \label{fig2}
\end{figure*}

\subsection{Transformer on 3D Object Detection}
Transformers have achieved significant success in both NLP \cite{attention} and image processing \cite{2020image, swin}, inspiring researchers to apply attention mechanisms to LiDAR-based 3D object detection. CT3D \cite{cwformer} introduces a channel-wise transformer to refine proposal bounding boxes. SST \cite{sst} employs a single-stride architecture using transformers to capture long-range dependencies. VoxSet \cite{vsformer} processes input point clouds through a set-to-set translation strategy with transformers. DSVT \cite{dsvt} and FlatFormer \cite{flatformer} adopt similar voxel partitioning strategies, sorting voxels within each window before grouping them into equal sizes. More recently, LION \cite{lion} applies linear attention to process grouped voxels within windows, while ScatterFormer \cite{scatterformer} uses linear attention to handle voxels of varied lengths within each window.

\subsection{Label Assignment}
Label assignment is a key component of 2D and 3D object detection, critically impacting network optimization. In 2D detectors, RetinaNet \cite{focalloss} assigns anchors to the output grid map, FCOS \cite{fcos} marks grid points within ground truth boxes as positive samples, and CenterNet \cite{centernet} selects the center points of ground truth boxes as positive samples. ATSS \cite{atss} and AutoAssign \cite{zhu2020autoassign} propose adaptive strategies for dynamic threshold selection and adjusting positive/negative confidence. For 3D dense detectors, label assignment typically follows anchor-based \cite{second,pointpillars,voxel_rcnn} or center-based \cite{centerpoint,afdet,afdetv2} approaches. Existing 3D sparse detectors \cite{voxelnext, safdnet} generally assign the nearest voxel to each object center as the positive sample. Recently, a 3D dense detector DCDet \cite{dcdet} dynamically assigns positive samples from a cross-shaped region around each object center. Inspired by this, we dynamically select positive samples from candidate sparse voxels near object centers.

\section{Methodology}

\subsection{Fully Sparse Hybrid Network}

Sparse detectors \cite{fsd, fsdv2, voxelnext, safdnet} extract features solely from sparse voxels, eliminating the need for dense feature maps. This makes them more efficient for long-range detection tasks compared to dense detectors \cite{second, centerpoint, dcdet, pv-rcnn}. However, existing sparse detectors face challenges in establishing long-range interactions between distant voxels. As shown in Figure~\ref{fig1}, for the sparse detector, no interactions occur between distant voxels if they fall outside each other's receptive fields. In contrast, the dense detector has intermediate empty voxels that can act as bridges, facilitating interactions between distant voxels.


Increasing the kernel size can expand the receptive field, which benefits 3D object detection \cite{largekernel3d, link}. However, naively enlarging the kernel size incurs significant time and space overhead. Some large kernel methods \cite{largekernel3d, link} are difficult to integrate into existing sparse detectors and struggle to establish the long-range interactions necessary for perceiving the global scene. Attention mechanisms \cite{attention, linear_attn} in vision Transformers \cite{2020image, han2023flatten} are known for their ability to facilitate long-range interactions. Recently, several studies \cite{dsvt, flatformer, scatterformer, lion} have used Transformers to extract features from raw point clouds. However, processing raw point clouds can be highly time- and resource-intensive due to their large size. In this work, we combine the sparse convolution encoder \cite{voxelnext, safdnet} with attention mechanisms, leveraging the strengths of both. The sparse convolution encoder efficiently extracts downsampled sparse voxel features, while the attention mechanism enables long-range interactions among these downsampled voxels.


The overall framework of our fully sparse hybrid network is illustrated in Figure~\ref{fig2}. It begins by converting the input point clouds into sparse voxels using a voxel feature encoding (VFE) module, following previous methods \cite{dsvt, safdnet}. A sparse convolution encoder \cite{voxelnext, safdnet} then extracts features from the sparse voxels and progressively downsamples them. Next, we use SlotFormer to establish long-range interactions among these downsampled voxels. The sparse voxels then pass through a sparse upsampling module to generate a more fine-grained representation. Finally, the upsampled sparse voxels are processed by the dynamic sparse head, which dynamically assigns positive samples for network optimization. The details of SlotFormer, sparse upsampling, and dynamic sparse head will be discussed in Sec.~\ref{slotformer}, Sec.~\ref{sparse_upsampling} and Sec.~\ref{dynamic_sparse_head}, respectively.

\subsection{SlotFormer Architecture}
\label{slotformer}

Existing transformer-based 3D detectors \cite{dsvt, flatformer, scatterformer, lion} use a window partition strategy to split voxels, enabling them to achieve a large receptive field. However, they are still limited by the window size. To further expand the receptive field, we propose SlotFormer, which facilitates long-range interactions at the global scene level. We divide the scene into several slots along the X or Y axes, with each slot covering the entire scene along one axis, as illustrated in Figure~\ref{fig2}. The primary challenge then lies in processing the voxels within each slot. Due to the large size of the slots and the sparse distribution of point clouds, the number of non-empty voxels in each slot can be large and vary greatly. Traditional self-attention has a computational complexity that grows quadratically with respect to input size, and it struggles to parallelize computations for inputs of varying lengths. As a result, self-attention is not the ideal solution for our approach.


To reduce computational costs and handle varying input lengths, we employ linear attention \cite{performer, efficientvit, han2023flatten, linear_attn} to process non-empty voxels within slots. Linear attention modifies the computation order of $Q$, $K$, and $V$ compared to traditional self-attention. It first multiplies the matrices of $K$ and $V$, and then multiplies the result by $Q$, thus avoiding the explicit computation of the attention matrix in self-attention. As a result, the computational complexity of linear attention is reduced from $\mathcal{O}\left(N^{2}\right)$ to $\mathcal{O}\left(N\right)$.

Specifically, given $N$ non-empty voxels $\mathcal{V}=\left\{\nu_{i} \mid \nu_{i}=\left[\left(x_{i}, y_{i}\right) ; f_{i}\right]\right\}_{i=1}^{N}$, where $\left(x_{i}, y_{i}\right) \in \mathbb{R} ^{2}$ and $f_{i} \in \mathbb{R} ^{1\times c}$ are the coordinate and feature of the voxel $\nu_{i}$, respectively. We first partition these voxels into several slots with the size of $s \times w$, where $s$ and $w$ are the size of the point cloud scene and the width of the slot, respectively: 

\begin{equation}
    \begin{aligned}
    d^{x}_{i} = \lfloor \frac{y_i}{w} \rfloor, d^{y}_{i} = \lfloor \frac{x_i}{w} \rfloor, 
    \end{aligned}
    \label{Eq:eq1}
\end{equation}

\noindent where $d^{x}_i$ and $d^{y}_i$ are the X-axis and Y-axis slot indices of voxel $\nu_{i}$, respectively. Next, we will demonstrate the calculation steps of the X-axis SlotFormer layer, therefore only $d^{x}_i$ is used. Query matrix $Q=\left\{q_{i} \in \mathbb{R}^{1\times c}\right\}_{i=1}^{N}$, key matrix $K=\left\{k_{i} \in \mathbb{R}^{1\times c}\right\}_{i=1}^{N}$ and value matrix $V=\left\{v_{i} \in \mathbb{R}^{1\times c}\right\}_{i=1}^{N}$ are obtained as follows:

\begin{equation}
    \begin{aligned}
    q_i = \phi (f_{i}W_q), k_i = \phi(f_{i}W_k), v_i = f_{i}W_v,
    \end{aligned}
    \label{Eq:eq2}
\end{equation}

\noindent where $W_q, W_k$ and $W_v$ are learnable weight matrics, and $\phi(\cdot)$ denotes the kernel function implemented by ReLU function. Next, we calculate the $KV=\left\{kv_{i} \in \mathbb{R}^{c \times c}\right\}_{i=1}^{m}$ matrix, where $m$ denotes the slot number:

\begin{equation}
    \begin{aligned}
    kv_j = \sum_{i=0}^{N} k_i^{T} \cdot v_i \cdot \mathbb{I}[d_i^x == j],
    \end{aligned}
    \label{Eq:eq3}
\end{equation}

\noindent where $\mathbb{I}[d_i^x == j]$ denotes whether the $i$-th voxel belongs to the $j$-th slot. The Eq.~\ref{Eq:eq3} sums up the products of $k_{i}^{T}$ and $v_{i}$ within a same slot. Here, we use the X-axis slots as an example. The process for Y-axis slots is similar to Eq.~\ref{Eq:eq3}. Now, each voxel can query its new value matrix $v_i^{\prime}$ as follows:

\begin{equation}
    \begin{aligned}
    v_{i}^{\prime} = \frac{q_i \cdot kv_{d_i^{x}}}{q_i \sum_{j=1}^{N} k^{T}_{j} \cdot \mathbb{I}[d_j^x == d_i^x]},
    \end{aligned}
    \label{Eq:eq4}
\end{equation}

\noindent where $kv_{d_i^{x}}$ denotes the $kv$ matrix of the $d_i^{x}$-th slot. The denominator of Eq.~\ref{Eq:eq4} is a normalization factor. Note that the above processes can be implemented with customized CUDA operations in parallel computation. Finally, the new value matrix $V^{\prime}=\left\{v_{i}^{\prime} \in \mathbb{R}^{1 \times c}\right\}_{i=1}^{N}$ will go through a feed-forward network:

\begin{equation}
    \begin{aligned}
    f_{i}^{\prime} = \text{FFN}(v_{i}^{\prime}).
    \end{aligned}
    \label{Eq:eq5}
\end{equation}

\noindent The updated voxels, $\mathcal{V}=\left\{\nu_{i} \mid \nu_{i}=\left[\left(x_{i}, y_{i}\right) ; f_{i}^{\prime}\right]\right\}_{i=1}^{N}$, will be input into the next layer. In the above case, we omit the residual connection and output projection in linear attention for simplicity.

\subsection{Sparse Upsampling}
\label{sparse_upsampling}

Current 3D sparse encoders typically employ a multi-stride downsampling strategy, which, while reducing computational cost, often leads to the loss of fine-grained information critical for small objects. To recover the lost details, we introduce a simple yet effective module called sparse upsampling.

Let's begin by considering $N$ non-empty voxels, $\mathcal{V}=\left\{\nu_{i} \mid \nu_{i}=\left[\left(x_{i}, y_{i}\right) ; f_{i}\right]\right\}_{i=1}^{N}$, where $\left(x_{i}, y_{i}\right) \in \mathbb{R}^{2}$ and $f_{i} \in \mathbb{R}^{1\times c}$ represent the coordinates and features of voxel $\nu_{i}$, respectively. Initially, we double the coordinates of these voxels (equivalently reducing the voxel size within the same point cloud range):

\begin{equation}
    \begin{aligned}
    \mathcal{V}^{\prime} = \left\{\nu_{i}^{\prime} \mid \nu_{i}^{\prime}=\left[\left(2x_{i}, 2y_{i}\right) ; f_{i}\right]\right\}_{i=1}^{N}.
    \end{aligned}
    \label{Eq:eq6}
\end{equation}

\noindent Then we employ a sparse convolution layer $\text{SpConv}(\cdot)$ with 3 kernel size and 1 stride to diffuse the voxels:


\begin{equation}
    \begin{aligned}
    \mathcal{V}^{up} =& \text{SpConv} (\mathcal{V}^{\prime}) \\
                     =&  \left\{\nu_{i}^{up} \mid \nu_{i}^{up}=\left[\left(x_{i}^{up}, y_{i}^{up}\right) ; f_{i}^{up}\right]\right\}_{i=1}^{N^{up}},
    \end{aligned}
    \label{Eq:eq7}
\end{equation}

\noindent where $\left(x_{i}^{up}, y_{i}^{up}\right)$ and $f_{i}^{up}$ denote the coordinates and features of the upsampled voxel $\nu_{i}^{up}$, respectively, and $N^{up}$ represents the total number of upsampled voxels. The sparse upsampling module refines the voxel size to one-fourth of the original size and leverages neighboring features to recover lost detailed information. 

\begin{figure}[t]
    \centering
    \includegraphics[width=0.9\columnwidth]{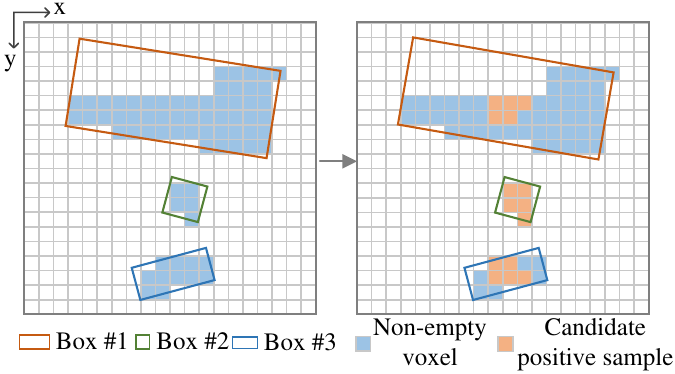}
    \caption{A demonstration of candidate positive samples in dynamic sparse label assignment. }
    \label{fig3}
\end{figure}

\begin{table*}[t]
    \centering
    \begin{tabular}{ccccccccc}
    \hline
    \multirow{2}{*}{Method}  & {LEVEL 2} & \multicolumn{3}{c}{LEVEL 1} && \multicolumn{3}{c}{LEVEL 2} \\
    \cline{3-5} \cline{7-9}
    & mAP/mAPH & Vehicle & Pedestrian & Cyclist && Vehicle & Pedestrian & Cyclist\\
    \hline
    SECOND \cite{second} & 61.0/57.2 & 72.3/71.7 & 68.7/58.2 & 60.6/59.3 && 63.9/63.3 & 60.7/51.3 & 58.3/57.0 \\
    PointPillars \cite{pointpillars} & 62.8/57.8 & 72.1/71.5 & 70.6/56.7 & 64.4/62.3 && 63.6/63.1 & 62.8/50.3 & 61.9/59.9 \\
    IA-SSD \cite{iassd} & 66.8/63.3 & 70.5/69.7 & 69.4/58.5 & 67.7/65.3 && 61.6/61.0 & 60.3/50.7 & 65.0/62.7 \\
    SST \cite{sst} & 67.8/64.6 & 74.2/73.8 & 78.7/69.6 & 70.7/69.6 && 65.5/65.1 & 70.0/61.7 & 68.0/66.9\\
    CenterPoint \cite{centerpoint} & 68.2/65.8 & 74.2/73.6 & 76.6/70.5 & 72.3/71.1 && 66.2/65.7 & 68.8/63.2 & 69.7/68.5 \\
    Voxel R-CNN \cite{voxel_rcnn} & 68.6/66.2 & 76.1/75.7 & 78.2/72.0 & 70.8/69.7 && 68.2/67.7 & 69.3/63.6 & 68.3/67.2 \\
    PV-RCNN \cite{pv-rcnn} & 69.6/67.2 & 78.0/77.5 & 79.2/73.0 & 71.5/70.3 &&  69.4/69.0 & 70.4/64.7 & 69.0/67.8 \\
    CenterFormer \cite{centerformer} & 71.1/68.9 & 75.0/74.4 & 78.6/73.0 & 72.3/71.3 &&  69.9/69.4 & 73.6/68.3 & 69.8/68.8 \\
    PillarNeXt \cite{pillarnext} & 71.9/69.7 & 78.4/77.9 & 82.5/77.1 & 73.2/72.2 && 70.3/69.8 & 74.9/69.8 & 70.6/69.6\\
    DSVT (Voxel) \cite{dsvt} & 74.0/72.1 & 79.7/79.3 & 83.7/78.9 & 77.5/76.5 && 71.4/71.0 & 76.1/71.5 & 74.6/73.7 \\
    DCDet \cite{dcdet} & 75.0/72.7 & 79.5/79.0 & 84.1/78.5 & 79.4/78.3 && 71.6/71.1 & 76.7/71.3 & 76.8/75.7\\
    LION-Mamba \cite{lion} & 75.1/73.2 & 79.5/79.1 & 84.9/80.4 & 79.7/78.7 && 71.1/70.7 & 77.5/73.2 & 76.7/75.8\\
    HEDNet \cite{hednet} & 75.3/73.4 & 81.1/80.6 & 84.4/80.0 & 78.7/77.7 && 73.2/72.7 & 76.8/72.6 & 75.8/74.9 \\
    ScatterFormer \cite{scatterformer} & 75.7/73.8 & 81.0/80.5 & 84.5/79.9 & 79.9/78.9 && 73.1/72.7 & 77.0/72.6 & 77.1/76.1 \\
    \hline
    SWFormer \cite{swformer}$^\dagger$ & -/- & 77.8/77.3 & 80.9/72.7& -/- && 69.2/68.8 & 72.5/64.9 & -/- \\
    VoxelNeXt \cite{voxelnext}$^\dagger$ & 72.2/70.1 & 78.2/77.7 & 81.5/76.3 & 76.1/74.9 && 69.9/69.4 & 73.5/68.6 & 73.3/72.2 \\
    FSDv1 \cite{fsd}$^\dagger$ & 72.9/70.8 & 79.2/78.8 & 82.6/77.3 & 77.1/76.0 && 70.5/70.1 & 73.9/69.1 & 74.4/73.3 \\
    FSDv2 \cite{fsdv2}$^\dagger$ & 75.6/73.5 & 79.8/79.3 & 84.8/79.7 & 80.7/79.6 && 71.4/71.0 & 77.4/72.5 & 77.9/76.8 \\
    SAFDNet \cite{safdnet}$^\dagger$ & 75.7/73.9 & 80.6/80.1 & 84.7/80.4 & 80.0/79.0 && 72.7/72.3 & 77.3/73.1 & 77.2/76.2 \\
    \rowcolor{cyan!20} FSHNet$_{light}$ (Ours)$^\dagger$ & 76.3/74.2 & 81.0/80.5 & 85.6/80.7 & 80.2/79.1 && 73.0/72.5 & 78.6/73.7 & 77.4/76.4\\
    \rowcolor{cyan!20} FSHNet$_{base}$ (Ours)$^\dagger$ & 77.1/74.9 & 82.2/81.7 & 85.9/80.8 & 80.5/79.4 && 74.5/74.0 & 78.9/73.9 & 78.0/76.9\\
    \hline
    \end{tabular}
    \caption{Performance comparisons on the Waymo Open \textit{validation} set. The results of AP/APH are reported. $\dagger$ represents a fully sparse detector, the same as below. All models are trained under the single-frame setting.}
\label{table1}
\end{table*}

\begin{table*}[t]
    \centering
    \begin{tabular}{ccccccccc}
    \hline
    \multirow{2}{*}{Method} & {LEVEL 2} & \multicolumn{3}{c}{LEVEL 1} && \multicolumn{3}{c}{LEVEL 2} \\
    \cline{3-5} \cline{7-9}
    & mAP/mAPH & Vehicle & Pedestrian & Cyclist && Vehicle & Pedestrian & Cyclist\\
    \hline
    CenterPoint \cite{centerpoint} & - & 80.2/79.7 & 78.3/72.1 & - && 72.2/71.8 & 72.2/66.4 & - \\
    AFDetV2 \cite{afdetv2} & 72.2/70.0 & 80.5/80.0 & 79.8/74.4 & 72.4/71.2 && 73.0/72.6 & 73.7/68.6 & 69.8/68.7\\
    PV-RCNN++ \cite{pv-rcnn++} & 72.4/70.2 & 81.6/81.2 & 80.4/75.0 & 71.9/70.8 && 73.9/73.5 & 74.1/69.0 & 69.3/68.2 \\
    FSDv2 \cite{fsdv2}$^\dagger$ & 75.4/73.3 & 82.4/82.0 & 83.8/78.9 & 77.1/76.0 && 74.4/74.0 & 77.4/72.8 & 74.3/73.2\\
    SAFDNet \cite{safdnet}$^\dagger$ & 76.5/74.6 & 83.9/83.5 & 84.3/79.8 & 77.5/76.3 && 76.6/76.2 & 78.4/74.1 & 74.6/73.4  \\
    \rowcolor{cyan!20} FSHNet$_{light}$ (Ours)$^\dagger$ & 77.0/74.9 & 83.4/83.0 & 84.7/79.8 & 78.9/77.7 && 76.1/75.7 & 78.8/74.1 & 76.1/74.9\\
    \rowcolor{cyan!20} FSHNet$_{base}$ (Ours)$^\dagger$ & 77.4/75.2 & 84.9/84.5 & 85.3/80.0 & 77.7/76.5 && 77.8/77.4 & 79.5/74.3 & 74.8/73.7\\
    \hline
    \end{tabular}
    \caption{Performance comparisons on the Waymo Open \textit{test} set by submitting to the official test evaluation server. The results are achieved by using the single-frame setting. No test-time augmentations are used. }
\label{table2}
\end{table*}

\subsection{Dynamic Sparse Head}
\label{dynamic_sparse_head}
\textbf{Dynamic Sparse Label Assignment. }
Existing sparse detectors \cite{voxelnext, safdnet} typically assign the voxel closest to the center of each annotated bounding box as a positive sample (hereafter referred to as center nearest assignment). However, this approach excludes many high-quality samples, leading to a suboptimal detection network. In contrast, the recent dense 3D detector DCDet \cite{dcdet} introduces a dynamic cross-based label assignment method that identifies positive samples within a cross-shaped area around each bounding box. Inspired by this, we propose a dynamic sparse label assignment strategy that selects positive samples from a set of the nearest candidate voxels to each bounding box center. This method aims to enhance network optimization by incorporating more high-quality positive samples during training. A visual demonstration of the dynamic sparse label assignment is shown in Figure~\ref{fig3}.

Specifically, the dynamic sparse label assignment first calculates the distances between voxels and the centers of bounding boxes. For each annotated bounding box $\textbf{b}^{t}$, we find the nearest $n$ voxels $\mathcal{V}_{b} =\left\{\nu_{i} \mid \nu_{i}=\left[\left(x_{i}, y_{i}\right) ; f_{i}\right]\right\}_{i=1}^{n}$ to its center. And then we calculate the selection cost $\mathcal{C}= \left\{c_{i} \right\}_{i=1}^{n}$ of these voxels as follows:


\begin{equation}
    c_{i} = \ell_{cls}(\text{Pred}(\nu_{i}), \textbf{b}^t) + \lambda \ell_{reg}(\text{Pred}(\nu_{i}), \textbf{b}^t), \nu_{i} \in \mathcal{V}_b,
\label{Eq:eq8}
\end{equation}

\noindent where $\text{Pred}(\cdot)$ denotes the prediction module which outputs the predicted bounding boxes according to input voxels, $\ell_{cls}(\cdot, \cdot)$ and $\ell_{reg}(\cdot, \cdot)$ are the classification and regression loss functions respectively, and $\lambda$ is the weight of regression loss. Then, we sum the IoUs between the ground truth $\mathbf{b}^{t}_{i}$ and predicted bounding boxes from the candidate voxels:

\begin{equation}
    k = \max(\lfloor \sum_{\nu_i \in \mathcal{V}_b} \mathrm{IoU}(\mathbf{b}^{t}, \text{Pred}(\nu_{i}))\rfloor, 1).
\label{Eq:eq9}
\end{equation}

\noindent Finally, top $k$ voxels in $\mathcal{V}_b$ are selected as positive samples for ground truth $\mathbf{b}^{t}$ according the selection cost $\mathcal{C}$. 


\noindent \textbf{Detection Loss. } 
We simply use two submanifold sparse convolutional layers with kernel size 1 to build the prediction module, without other complicated designs. For regression, predicted bounding boxes are generated based on the features of positive samples obtained by the dynamic sparse label assignment. We utilize the rotation-weighted IoU loss \cite{dcdet} as the regression loss. For classification, we adopt the heatmap target strategy. The weights of positive samples are set to 1, and the weights of negative samples from candidate voxels are set to the values of IoU between predicted and ground-truth boxes. As for the rest negative samples, the weights are all set to 0. We apply the focal loss \cite{focalloss} as the classification loss.  

\section{Experiments}

\begin{table*}[t]
    \centering
    \begin{tabular}{lcccccccccccc}
        \toprule
        Method & NDS & mAP & Car & Truck & Bus & T.L. & C.V. & Ped. & M.T. & Bike & T.C. & B.R. \\
        \midrule
        CenterPoint~\cite{centerpoint} & 66.5 & 59.2 & 84.9 & 57.4 & 70.7 & 38.1 & 16.9 & 85.1 & 59.0 & 42.0 & 69.8 & 68.3 \\
        PillarNeXt~\cite{pillarnext} & 68.4 & 62.2 & 85.0 & 57.4 & 67.6 & 35.6 & 20.6 & 86.8 & 68.6 & 53.1 & 77.3 & 69.7 \\
        VoxelNeXt~\cite{voxelnext}$\dagger$ & 68.7 & 63.5 & 83.9 & 55.5 & 70.5 & 38.1 & 21.1 & 84.6 & 62.8 & 50.0 & 69.4 & 69.4 \\
        TransFusion-L\cite{transfusion} & 70.1 & 65.5 & 86.9 & 60.8 & 73.1 & 43.4 & 25.2 & 87.5 & 72.9 & 57.3 & 77.2 &  70.3\\
        FSDv2~\cite{fsdv2}$^\dagger$ & 70.4 & 64.7 & 84.4 & 57.3 & 75.9 & 44.1 & 28.5 & 86.9 & 69.5 & 57.4 & 72.9 & 73.6 \\
        SAFDNet \cite{safdnet}$^\dagger$ & 71.0 & 66.3 & 87.6 & 60.8 & 78.0 & 43.5 & 26.6 & 87.8 & 75.5 & 58.0 & 75.0 & 69.7 \\
        DSVT~\cite{dsvt} & 71.1 & 66.4 & 87.4 & 62.6 & 75.9 & 42.1 & 25.3 & 88.2 & 74.8 & 58.7 & 77.8 & 70.9 \\
        \rowcolor{cyan!20} FSHNet$_{base}$ (Ours)$^\dagger$ & 71.7 & 68.1 & 88.7 & 61.4 & 79.3 & 47.8 & 26.3 & 89.3 & 76.7 & 60.5 & 78.6 & 72.3 \\
        \bottomrule
    \end{tabular}
    \caption{Performance comparisons on the nuScenes \textit{validation} set. `T.L.', `C.V.', `Ped.', `M.T.', `T.C.', and `B.R.' denote trailer, construction vehicle, pedestrian, motor, traffic cone, and barrier, respectively.}
\label{table3}    
\end{table*}

\begin{table*}[t]
   \centering
   \scriptsize
   \resizebox{\linewidth}{!}{
   \small
    \begin{tabular}{c|rrrrrrrrrrrrrrrrrrrrrrrrrrr}
    \hline
    Method & \rotatebox{90}{mAP} &\rotatebox{90}{Vehicle} & \rotatebox{90}{Bus} & \rotatebox{90}{Pedestrian} & \rotatebox{90}{Stop Sign} & \rotatebox{90}{Box Truck} & \rotatebox{90}{Bollard} & \rotatebox{90}{C-Barrel} & \rotatebox{90}{Motorcyclist} & \rotatebox{90}{MPC-Sign} & \rotatebox{90}{Motorcycle} &\rotatebox{90}{Bicycle} & \rotatebox{90}{A-Bus} & \rotatebox{90}{School Bus} & \rotatebox{90}{Truck Cab} & \rotatebox{90}{C-Cone} & \rotatebox{90}{V-Trailer} & \rotatebox{90}{Sign} & \rotatebox{90}{Large Vehicle} & \rotatebox{90}{Stroller} & \rotatebox{90}{Bicyclist} \\
    \hline
    CenterPoint \cite{centerpoint} & 22.0 & 67.6 & 38.9 & 46.5 & 16.9 & 37.4 & 40.1 & 32.2 & 28.6 & 27.4 & 33.4 & 24.5 & 8.7 & 25.8 & 22.6 & 29.5 & 22.4 & 6.3 & 3.9 & 0.5 & 20.1 \\
    HEDNet \cite{hednet} & 37.1 & 78.2 & 47.7 & 67.6 & 46.4 & 45.9 & 56.9 & 67.0 & 48.7 & 46.5 & 58.2 & 47.5 & 23.3 & 40.9 & 27.5 & 46.8 & 27.9 & 20.6 & 6.9 & 27.2 & 38.7 \\
    FSDv1 \cite{fsd}$\dagger$ & 28.2 & 68.1 & 40.9 & 59.0 & 29.0 & 38.5 & 41.8 & 42.6 & 39.7 & 26.2 & 49.0 & 38.6 & 20.4 & 30.5 & 14.8 & 41.2 & 26.9 & 11.9 & 5.9 & 13.8 & 33.4 \\
    FSDv2 \cite{fsdv2}$\dagger$ & 37.6 & 77.0 & 47.6 & 70.5 & 43.6 & 41.5 & 53.9 & 58.5 & 56.8 & 39.0 & 60.7 & 49.4 & 28.4 & 41.9 & 30.2 & 44.9 & 33.4 & 16.6 & 7.3 & 32.5 & 45.9 \\
    VoxelNeXt \cite{voxelnext}$^{*}\dagger$& 33.1 & 74.8 & 43.2 & 65.5 & 36.3 & 43.1 & 58.7 & 67.5 & 29.4 & 43.3 & 54.0 & 45.4 & 21.7 & 41.8 & 23.8 & 51.2 & 28.3 & 17.9 & 6.3 & 13.1 & 29.8 \\
    SAFDNet \cite{safdnet}$^{*}\dagger$ & 38.7& 77.7 & 47.3 & 69.4 & 49.8 & 45.9 & 62.2 & 69.5 & 52.5 & 45.4 & 59.5 & 48.5 & 29.6 & 43.3 & 25.7 &  54.2 & 31.2 & 20.5 & 9.3 & 36.0 & 48.5 \\
    \rowcolor{cyan!20} FSHNet$_{base}$ (Ours)$\dagger$& 40.2 & 77.4 & 48.3 & 72.9 & 50.0 & 47.2 & 63.4 & 69.9 & 56.1 & 43.8 & 62.5 & 53.7 & 31.8 & 44.8 & 28.9 &  56.6 & 32.0 & 23.0 & 8.7 & 34.5 & 41.0 \\    
    \hline
    \end{tabular}}
   \caption{Performance comparisons on the Argoverse2 \textit{validation} set. `*' represents that the method is re-implemented using the codebase OpenPCDet \cite{openpcdet}. We exclude categories with few instances but still calculate the average across all categories, including those omitted.}
\label{table4}
\end{table*}

\subsection{Dataset} 
\noindent\textbf{Waymo Open. }We perform our primary experiments on the Waymo Open dataset \cite{waymo}. It is a large and widely-used benchmark for LiDAR-based 3D object detection. The Waymo Open dataset comprises 1150 sequences, totaling over 200,000 frames, with 798 sequences for training, 202 for validation, and 150 for testing. 

\noindent\textbf{nuScenes. }We also conduct experiments on the nuScenes dataset \cite{nuscenes} to validate the generalization of our FSHNet. The dataset includes 1000 driving scenarios, with 700 scenarios for training, 150 for validation, and 150 for testing. 

\noindent\textbf{Argoverse2. } Furthermore, we conduct experiments on the Argoverse2 (AV2) dataset \cite{argo2} to validate the effectiveness of FSHNet in long-range detection. AV2 dataset encompasses 1000 driving scenarios, divided into 700 for training, 150 for validation, and 150 for testing. Compared to the Waymo Open and nuScenes datasets, AV2 features a greater number of long-range 3D cuboid annotations. 

\subsection{Implementation Details}
Our method is implemented using the OpenPCDet framework \cite{openpcdet}. All models are trained from scratch in an end-to-end fashion with the Adam optimizer and a learning rate of 0.003. There are two variants of FSHNet: one integrates the sparse encoder of VoxelNeXt with 4 SlotFormer layers, termed FSHNet$_{\textit{light}}$; the other integrates the sparse encoder of SAFDNet with 8 SlotFormer layers, termed FSHNet$_{\textit{base}}$. The slot width $w$ for SlotFormer is set to 12. We set the number of candidate voxels in sparse dynamic label assignment to 5. We observe that our FSHNet achieves convergence more rapidly than previous methods. Consequently, the FSHNet variants are trained for 12, 36, and 12 epochs on the Waymo Open, nuScenes (without the CBGS strategy \cite{cbgs}), and Argoverse2 datasets, respectively. The detection ranges of the three datasets are 75, 54, and 200 meters, respectively. The voxel sizes are $(0.08, 0.08, 0.15)m$, $(0.075, 0.075, 0.2)m$, and $(0.1, 0.1, 0.2) m$ for the three datasets, respectively. All experiments are conducted on 2 A100 GPUs with a total batch size of 16. Additional experiments including visualizations, runtime, and hyper-parameter analysis are in supplementary materials. 

\subsection{Comparison with State-of-the-Art Methods}
\vspace{-0.5em}
\noindent\textbf{Results on Waymo Open.}
We conduct a comprehensive comparison of FSHNet against the current state-of-the-art (SOTA) 3D detectors on the validation set of the Waymo Open Dataset. As demonstrated in Table \ref{table1}, our FSHNet$_{\textit{light}}$ and FSHNet$_{\textit{base}}$ outperform most dense and sparse detectors, including the most recent SOTA methods such as ScatterFormer \cite{scatterformer} and SAFDNet \cite{safdnet}. Moreover, FSHNet$_{\textit{base}}$ surpasses the previous methods by a substantial margin. Additionally, we evaluated our method on the Waymo Open test set by submitting the results to the official server. The performance comparisons are presented in Table \ref{table2}, revealing that our FSHNet$_{\textit{light}}$ and FSHNet$_{\textit{base}}$ achieve SOTA performance according to the LEVEL 2 mAP/mAPH. Notably, FSHNet$_{\textit{base}}$ significantly surpasses previous methods, particularly in the vehicle and pedestrian categories. The improvement mentioned above can be attributed to the components of FSHNet. SlotFormer enhances the long-range interaction capability, which is beneficial for detecting large objects. The sparse upsampling module captures fine-grained content for small objects. The sparse dynamic label assignment better optimizes the detection network.

\noindent\textbf{Results on nuScenes.}
We further conduct experiments on the nuScenes dataset to compare our method with other SOTA LiDAR-based methods. Our FSHNet$_\textit{base}$ are trained for 36 epochs without CBGS strategy \cite{cbgs}, which typically extends training iterations by about 4.5 times. Compared to other methods trained for 20 epochs with CBGS, our FSHNet$_\textit{base}$ undergoes fewer training iterations. Nevertheless, it still outperforms other methods in NDS and mAP metrics, as shown in Table~\ref{table3}.

\noindent\textbf{Results on Argoverse2.}
To validate the effectiveness of FSHNet on long-range detection, we conduct experiments on the Argoverse2 dataset with a perception range of 200 meters. We re-implement two recent sparse detectors including VoxelNeXt and SAFDNet for fair comparison. The re-implemented detectors adopt the same training scheme as our method, including ground-truth sampling to alleviate the long-tail problem. As demonstrated in Table~\ref{table4}, our FSHNet$_{\textit{base}}$ substantially outperforms the current SOTA sparse detector SAFDNet, across small objects (pedestrian +3.2\% mAP, bicycle +5.2\% mAP) and large objects (bus +1.0\% mAP, A-bus +2.2\% mAP).

\subsection{Ablation Study}


To further study the influence of each component of FSHNet, we perform a comprehensive ablation analysis on the Waymo Open dataset. Following prior works \cite{dsvt, dcdet}, models are trained on 20\% training samples and the LEVEL 2 AP/APH results on the whole \textit{validation} set are reported.

\noindent\textbf{Effect of Different Components.}
We employ VoxelNeXt with the center nearest label assignment as our baseline model. To assess the effectiveness of our proposed methods, we systematically integrate SlotFormer (\textit{SF}), dynamic sparse label assignment (\textit{DSLA}), and sparse upsampling (\textit{SU}) into the baseline model. The ablation study results are shown in Table~\ref{table5}. We note a significant performance enhancement when incorporating SlotFormer, as evidenced by the results in the 2$^{nd}$ and 6$^{th}$ rows of Table~\ref{table5}. Particularly for large objects such as vehicles and cyclist categories, the improvements are marked, illustrating the importance of long-range interactions. Additionally, models trained with \textit{DSLA} consistently yield significantly better performance, as portrayed in the 3$^{rd}$ and 5$^{th}$ rows of Table~\ref{table5}. This suggests that \textit{DSLA} aids in the optimization of the detection network. We also observe that the sparse upsampling significantly enhances the performance on small objects (e.g., pedestrians) as demonstrated in the 4$^{th}$ row of Table~\ref{table5}. This affirms the effectiveness of fine-grained details preserved by the sparse upsampling.

\begin{table}[t]
    \centering
    \begin{tabular}{cccccc}
        \hline
        \textit{SF} & \textit{DSLA}  & \textit{SU} & {Vehicle} & {Pedestrian} & {Cyclist}\\
        \hline
        &             &            & 69.1/68.7 & 75.3/69.5 & 75.0/73.9 \\ 
        $\checkmark$ & &            & 70.3/69.9 & 75.9/70.5 & 76.2/75.1 \\
        &$\checkmark$ & & 69.9/69.5 & 75.5/69.7 & 75.7/74.5 \\
        &&$\checkmark$ & 69.3/68.8 & 76.6/71.0 & 75.2/74.1 \\
        &$\checkmark$&$\checkmark$& 70.5/70.1 & 77.1/71.6& 75.6/74.5\\
        \rowcolor{lgray} $\checkmark$&$\checkmark$&$\checkmark$ & 72.5/72.0 & 77.9/72.6 & 77.2/76.1\\
        \hline
    \end{tabular}
    \caption{Effect of different components of FSHNet.}
\label{table5}
\end{table}

\begin{table}[t]
    \centering
    \resizebox{\linewidth}{!}{
    \begin{tabular}{cccc}
        \hline
        Voxel Partition & {Vehicle} & {Pedestrian} & {Cyclist}\\
        \hline
        baseline          & 70.5/70.1 & 77.1/71.6 & 75.6/74.5 \\ 
        Win.+Set w/ self-attn & 71.8/71.3 & 77.3/72.0 & 76.8/75.7\\
        Win.+Set w/ linear attn & 72.0/71.5 & 77.7/72.3 & 77.1/76.0\\ 
        Win. w/ linear attn & 72.2/71.7 & 77.8/72.4 & 77.1/76.1\\ 
        \rowcolor{lgray} Slot w/ linear attn & 72.5/72.0 & 77.9/72.6& 77.2/76.1\\ 
        \hline
    \end{tabular}}
    \caption{Comparison results of different voxel partition manners. }
\label{table6}
\end{table}

\noindent\textbf{Different Types of Voxel Partition Manners.}
Table~\ref{table6} presents a comparison of results across different voxel partition manners. We employ the FSHNet$_{light}$ without the SlotFormer as the baseline. `Win.+Set' refers to partitioning voxels into windows, then sorting and grouping them into equal-sized sets, as in \cite{dsvt}. `Win.' represents only assigning voxels to distinct windows, and `Slot' is our partition approach, detailed in Sec.~\ref{slotformer}. `w/ self-attn' and `w/ linear attn' indicate the use of self-attention and linear attention mechanisms for voxel feature extraction, respectively. As shown in Table~\ref{table6}, the attention mechanisms combined with different voxel partition manners significantly improve the baseline model’s performance, attributed to their long-range interaction capability. Among these manners, our slot partition with linear attention achieves the best results, as it effectively establishes global-level voxel interactions.

\noindent\textbf{Different Types of Upsampling Strategies.}
Table~\ref{table7} provides a comparison of various upsampling strategies for sparse voxels. We use FSHNet$_{light}$ without the sparse upsampling module as the baseline, shown in the $1^{st}$ row of Table~\ref{table7}. \textit{SM-SU} denotes a method where the features of large voxels are repeated to their inner small voxels, followed by a submanifold sparse convolution layer \cite{second} to process these features. \textit{SP-SU}, in contrast, assigns the features of large voxels to one inner small voxel, then uses a sparse convolution layer \cite{second} to diffuse the voxel features. Both the \textit{SM-SU} and \textit{SP-SU} upsampling strategies significantly improve detection performance on small objects (e.g., pedestrians) compared to the baseline model. Additionally, the \textit{SP-SU} upsampling strategy enhances performance on large objects (e.g., vehicles and cyclists), likely because its voxel diffusion mitigates the issue of center feature missing.

\begin{table}[t]
    \centering
    \begin{tabular}{cccc}
        \hline
        Type & {Vehicle} & {Pedestrian} & {Cyclist} \\
        \hline
        - & 71.4/70.9 & 75.9/70.4 & 76.6/75.4\\
        \textit{SM-SU}     & 71.7/71.2 & 77.9/72.4 & 76.6/75.6 \\
        \rowcolor{lgray} \textit{SP-SU}  & 72.5/72.0 & 77.9/72.6 & 77.2/76.1 \\
       
        \hline
    \end{tabular}
    \caption{Comparison results of different sparse upsampling types. }
\label{table7}
\end{table}

\begin{table}[t]
    \centering
    \begin{tabular}{cccc}
        \hline
        $n$ & {Vehicle} & {Pedestrian} & {Cyclist} \\
        \hline
        -     & 70.8/70.3 & 77.3/71.7 & 76.1/75.0 \\
        1  & 70.8/70.3 & 77.1/71.5 & 77.0/76.0 \\
        3  & 71.3/70.8 & 77.6/72.0 & 77.0/75.9 \\
        \rowcolor{lgray} 5  & 72.5/72.0 & 77.9/72.6 & 77.2/76.1 \\
        7  & 72.2/71.7 & 77.8/72.4 & 77.2/76.0 \\
        \hline
    \end{tabular}
    \caption{Comparison results of different candidate numbers. }
\label{table8}
\end{table}

\noindent\textbf{Different Candidate Number of \textit{DSLA}.}
Table~\ref{table8} demonstrates a comparison of different candidate numbers for \textit{DSLA}. We use FSHNet$_{light}$ without \textit{DSLA} (employing the center nearest assignment strategy) as the baseline, shown in the $1^{st}$ row of Table~\ref{table8}. Increasing the candidate number from 1 to 5 progressively enhances the detection performance, as demonstrated in the $2^{nd}$ to $4^{th}$ rows of Table~\ref{table8}. However, further increasing the candidate number does not yield additional improvement, as shown in the last row of Table~\ref{table8}. Therefore, we set the candidate number to 5 as the default configuration.

\section{Conclusion}
In this paper, we propose a sparse detector termed FSHNet which combines the efficiency of sparse convolutions with the long-range interaction capabilities of the attention mechanism within a single model. It incorporates the SlotFormer block to facilitate long-range interactions between sparse voxels. Additionally, FSHNet employs dynamic sparse label assignment to generate high-quality positive samples, enhancing network optimization. To further boost performance for small objects, we introduce a sparse upsampling module that preserves fine-grained details. Extensive experiments on several datasets demonstrate the effectiveness and generalization of FSHNet.

\noindent \textbf{Limitations.} We employ the SlotFormer to enable long-range interactions between sparse voxels. However, it introduces additional latency to sparse detectors. Specifically, for FSHNet$_{light}$ and FSHNet$_{base}$, SlotFormer adds approximately 16 \textit{ms} (from 65 to 81 \textit{ms}) and 29 ms (from 94 to 123 \textit{ms}) of latency, respectively, when tested on a single RTX 3090 GPU. A more efficient sparse architecture could potentially achieve the same long-range interaction capability as FSHNet. We leave this for future research.

\section*{Acknowledgments}
\noindent This work was supported in part by the National Natural Science Foundation of China under Grant 62232008, in part by Guangdong Basic and Applied Basic Research Foundation under Grant 2025A1515011485, in part by the National Natural Science Foundation of China under Grant 61902442, and in part by Guangdong Basic and Applied Basic Research Foundation under Grant 2025A1515010252.

{
    \small
    \bibliographystyle{ieeenat_fullname}
    \bibliography{main}
}

\clearpage

\appendix

\begin{figure*}[t]
    \centering
    \includegraphics[width=0.9\textwidth]{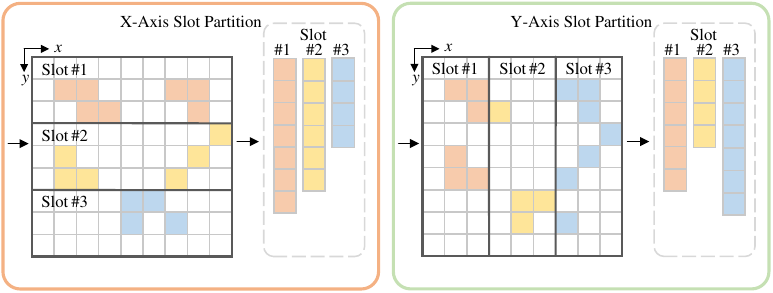}
    \caption{A demonstration of slot partition manner for sparse voxels. }
    \label{fig4}
\end{figure*}

\begin{figure*}[t]
    \centering
    \includegraphics[width=0.9\textwidth]{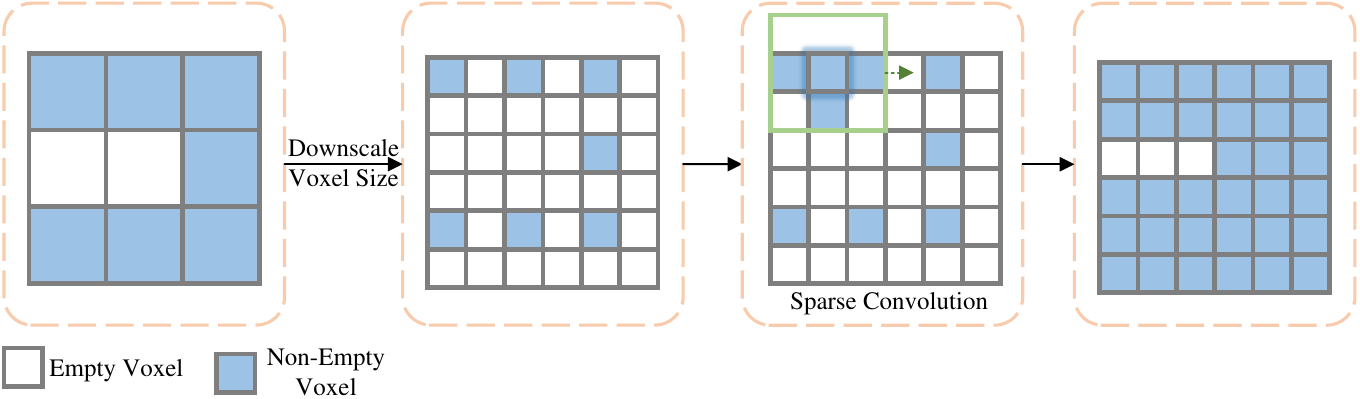}
    \caption{A demonstration of sparse voxel upsampling. }
    \label{fig5}
\end{figure*}

\begin{figure*}[t]
    \centering
    \includegraphics[width=0.9\textwidth]{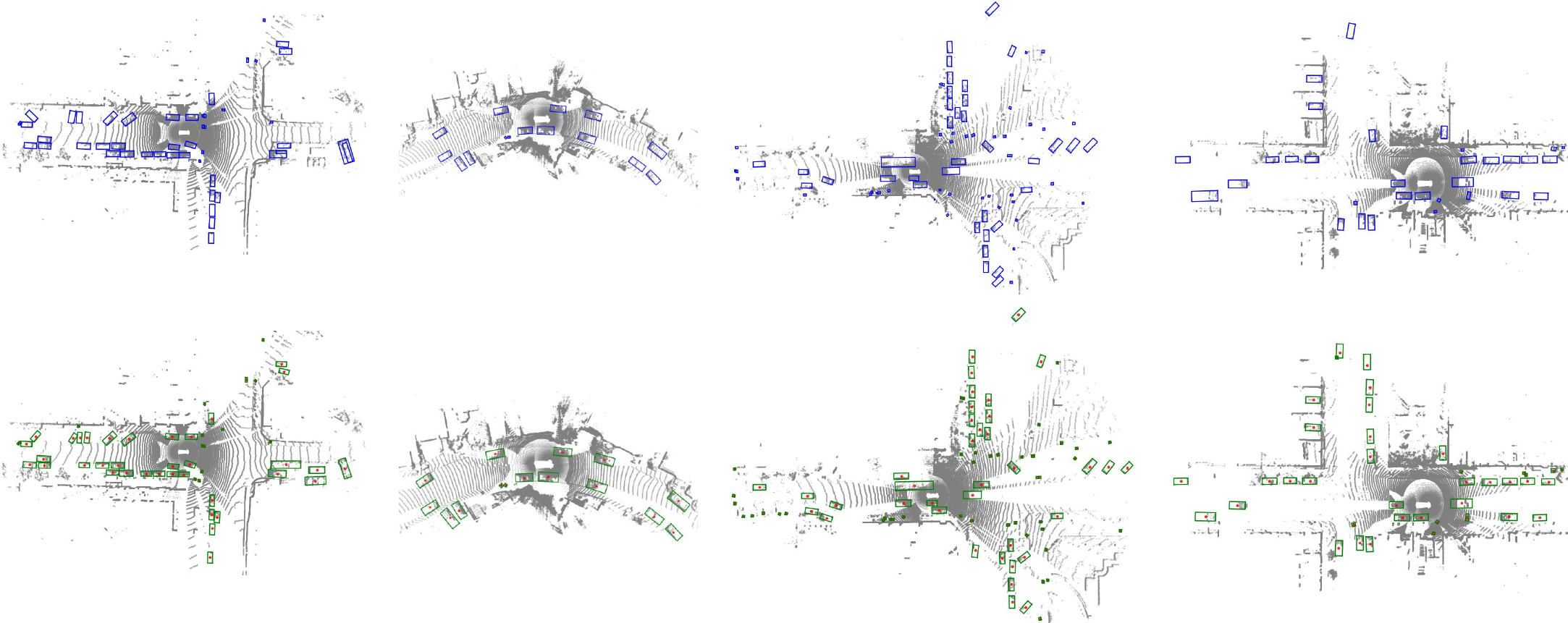}
    \caption{Qualitative results of FSHNet. The 1$^{st}$ row demonstrates the ground-truth boxes with blue color, and the 2$^{nd}$ row illustrates the predicted boxes with green color. }
    \label{fig6}
\end{figure*}

\begin{table}[t]
    \centering
    \resizebox{\linewidth}{!}{
    \begin{tabular}{ccccc}
        \hline
         Method & {Vehicle} & {Pedestrian} & {Cyclist} & Latency\\
        \hline
        \rowcolor{lgray} FSHNet$_\textit{light}$  & 72.5/72.0 & 77.9/72.6 & 77.2/76.1 & 81\\ 
        - \textit{SF}            & 70.5/70.1 & 77.1/71.6 & 75.6/74.5 & 65\\ 
        - \textit{SU}    & 71.4/70.9 & 75.9/70.4 & 76.6/75.4 & 71\\
        - \textit{DSLA}  & 70.8/70.3 & 77.8/72.2 & 76.1/75.0 & 81\\
        \hline
    \end{tabular}}
    \caption{Comparison of different FSHNet variants. The latency (\textit{ms}) is tested on a single RTX 3090 GPU. The LEVEL 2 AP/APH results (20\% training data) are reported. `-' denotes FSHNet$_\textit{light}$ without using corresponding module. }
    
\label{tableA1}
\end{table}

\begin{table}[t]
    \centering
    \resizebox{\linewidth}{!}{
    \begin{tabular}{ccccc}
        \hline
         & {Vehicle} & {Pedestrian} & {Cyclist} & Latency\\
        \hline
        VoxelNext \cite{voxelnext} & 69.9/69.4 & 73.5/68.6 & 73.3/72.2 & 56\\
        SAFDNet \cite{safdnet} & 72.7/72.3 & 77.3/73.1 & 77.2.76.2 & 94\\
        \rowcolor{lgray} FSHNet$_\textit{light}$  & 73.0/72.5 & 78.6/73.7 & 77.4/76.4 & 81 \\ 
        \rowcolor{lgray} FSHNet$_\textit{base}$  & 74.5/74.0 & 78.9/73.9 & 78.0/76.9 & 123 \\ 
        \hline
    \end{tabular}}
    \caption{Comparison of different sparse detectors. The latency (\textit{ms}) is tested on a single RTX 3090 GPU. The LEVEL 2 AP/APH results (100\% training data) are reported.}
\label{tableA2}
\end{table}

\begin{table}[t]
    \centering
    \begin{tabular}{ccccc}
        \hline
        $w$ & mAP/mAPH & {Vehicle} & {Pedestrian} & {Cyclist}\\
        \hline
        6       & 75.8/73.5 & 72.1/71.7 & 78.1/72.8 & 77.1/76.0 \\
        \rowcolor{lgray} 12      & 75.9/73.6 & 72.5/72.0 & 77.9/72.6 & 77.2/76.1 \\
        24      & 75.7/73.4 & 72.2/71.8 & 77.8/72.5 & 77.1/76.0 \\
        36      & 75.8/73.5 & 72.1/71.7 & 77.8/72.4 & 77.4/76.3 \\
        \hline
    \end{tabular}
    \caption{Effect of different $w$ settings. The LEVEL 2 AP/APH results (20\% training data) are reported.}
\label{tableA3}
\end{table}

\section{Runtime Analysis}
In this section, we first discuss the latency of different components of FSHNet and then compare the inference latency of various sparse detectors. Models are trained on the Waymo Open dataset. All latency measurements were conducted on a single RTX 3090 GPU. As shown in the 1$^{st}$ and 2$^{nd}$ rows of Table~\ref{tableA1}, the SlotFormer (\textit{SF}) block adds 16 \textit{ms} of latency to FSHNet$_{light}$, while significantly enhancing detection performance, particularly for large objects. As demonstrated in the 1$^{st}$ and 3$^{rd}$ rows of Table~\ref{tableA1}, the sparse upsampling (\textit{SU}) module introduces an additional 10 \textit{ms} latency to the detector, yet it markedly improves performance on small objects. As illustrated in the 1$^{st}$ and 4$^{th}$ rows of Table~\ref{tableA1}, our dynamic sparse label assignment (\textit{DSLA}) significantly boosts detection performance without adding latency.

We further compare the inference latency of our FSHNet with existing sparse detectors. As shown in Table~\ref{tableA2}, compared to the current state-of-the-art sparse detector SAFDNet, our FSHNet$_{light}$ exhibits lower inference latency and superior detection performance. Regarding our FSHNet$_{base}$, although it has relatively high inference latency, it greatly extends detection accuracy compared to existing sparse detectors.

\section{Hyper-Parameter Analysis}
In this section, we determine the optimal value for the slot width $w$ in Eq.\ref{Eq:eq1} through experiments on the Waymo Open dataset. The performance for different $w$ settings is shown in Table~\ref{tableA3}, indicating minimal variations. This is due to our SlotFormer having a global receptive field that is independent of slot width. When $w = 12$, there is a slightly better performance compared to other settings. Thus, we adopt $w = 12$ as the default setting.

\section{Visualizations}
To provide an intuitive understanding of the slot partitioning manner and the sparse upsampling strategy, we present visual demonstrations of each. As shown in Figure~\ref{fig4}, the slot partitioning process first scatters sparse voxels into grids and then groups them into different slots along the X- and Y-axes, respectively. Figure~\ref{fig5} illustrates the sparse upsampling strategy, where coarse voxels are initially compressed into smaller grids, followed by the application of sparse convolution to diffuse and refine them. Additionally, qualitative results are presented in Figure~\ref{fig6}, demonstrating our method’s ability to handle diverse and complex traffic scenes. The predicted boxes closely match the ground-truth boxes within an extensive detection range. However, we observe some missed or false detections in cases where objects are heavily occluded or located at extreme distances.
\end{document}